\documentclass{bmvc2k}

\def\eg{\textit{e.g.~}}
\def\ie{\textit{i.e.~}}

\def\etal{\textit{et~al.~}}

\usepackage{cleveref}
\crefname{subfigure}{fig.}{fig.}

\graphicspath{{figs/pdf/}}

\title{Depth Extraction from Videos Using Geometric Context and Occlusion Boundaries}

\addauthor{S. Hussain Raza}{hussain.raza@gatech.edu}{1}
\addauthor{Omar Javed}{omar.javed@sri.com}{2}
\addauthor{Aveek Das}{aveek.das@sri.com}{2}
\addauthor{Harpreet Sawhney}{harpreet.sawhney@sri.com}{2}
\addauthor{Hui Cheng}{hui.cheng@sri.com}{2}
\addauthor{Irfan Essa}{irfan@cc.gatech.edu}{3}

\addinstitution{
School of Electrical and Computer Engineering\\
 Georgia Institute of Technology\\
 Atlanta, Georgia, USA \\\vspace{0.15cm}
}
\addinstitution{
 SRI International\\
 Princeton, New Jersey, USA\\\vspace{0.15cm}
}
\addinstitution{
School of Interactive Computing\\
Georgia Institute of Technology\\
Atlanta, Georgia, USA\\\vspace{0.55cm}\hspace{-0.5cm}
 {{\footnotesize http://www.cc.gatech.edu/cpl/projects/videodepth}}
}
 
\runninghead{Raza et al.}{Depth Extraction from Videos}

\begin{document}

\maketitle

\begin{abstract}
\noindent We present an algorithm to estimate depth in dynamic video scenes. We propose to learn and infer depth in videos from appearance, motion, occlusion boundaries, and geometric context of the scene. Using our method, depth can be estimated from unconstrained videos with no requirement of camera pose estimation, and with significant background/foreground motions. We start by decomposing a video into spatio-temporal regions. For each spatio-temporal region, we learn the relationship of depth to visual appearance, motion, and geometric classes. Then we infer the depth information of new scenes using piecewise planar parametrization estimated within a Markov random field (MRF) framework by combining appearance to depth learned mappings and occlusion boundary guided smoothness constraints. Subsequently, we perform temporal smoothing to obtain temporally consistent depth maps. To evaluate our depth estimation algorithm, we provide a novel dataset with ground truth depth for outdoor video scenes. We present a thorough evaluation of our algorithm on our new dataset and the publicly available Make3d static image dataset.
\end{abstract}
\section{Introduction}
\label{sec:intro}

\begin{figure}
  \begin{minipage}[t]{0.47\linewidth}
  \centering
      \includegraphics[width=\textwidth]{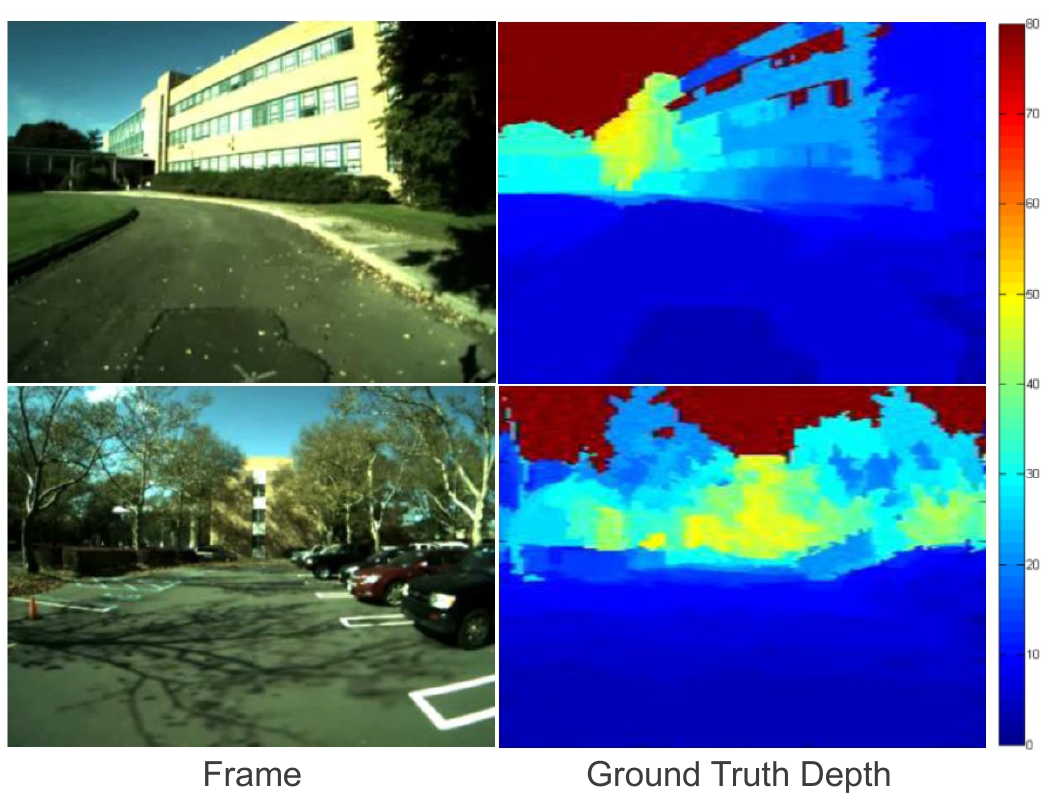}
  \caption{Video scene (left) and the ground truth depth from LiDAR (right).} 
  \label{fig:intro}
  \end{minipage}
  \hspace{0.5cm}
	\begin{minipage}[t]{0.47\linewidth}
	\centering
 \includegraphics[width=0.9\linewidth]{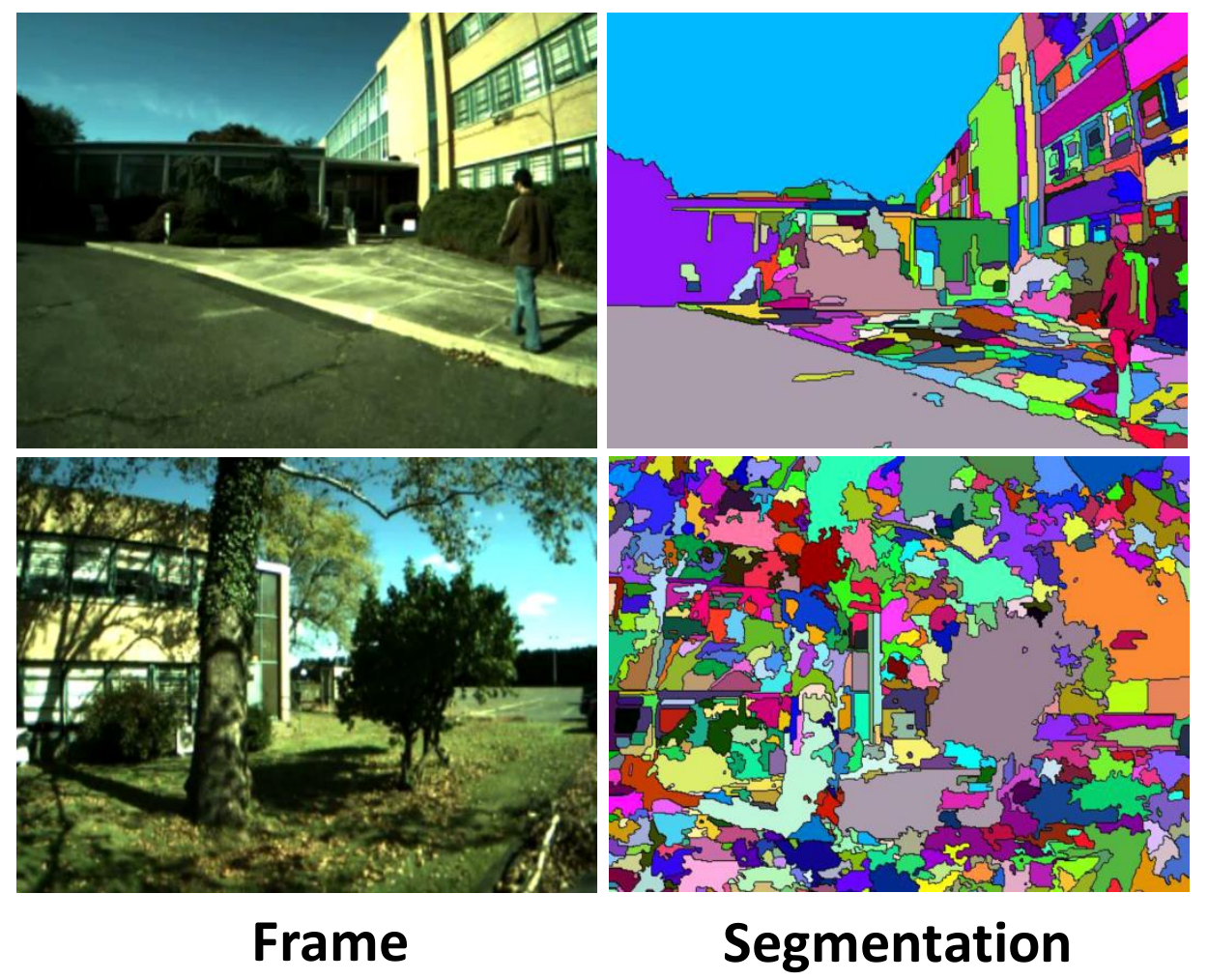}
  \caption{Video scenes and spatio-temporal segmentation using the method proposed by Grundmann \etal \cite{MatthiasSegmentation}.}
  \label{fig:vidSeg}
	\end{minipage}
\end{figure}

\noindent Human vision is capable of inferring the depth of a scene even when observed in monocular imagery. However, estimating depth from videos of general scenes still remains a challenging problem in computer vision.
A large majority of prior work on 3D structure extraction from a monocular camera, like structure from motion and multi-view stereo relies on pose estimation and  triangulation to estimate depth maps \cite{zhang2009consistent,newcombe2010live}(see Figure \ref{fig:intro}). In this paper, instead of using triangulation based constraints,  we propose to learn a mapping of monocular cues to depth using statistical learning and inference techniques. There are numerous monocular cues such as texture variations, occlusion boundaries,  defocus, color/haze, surface layout, and size/shape of known objects, that contain useful and important cues for depth information. These sources of depth information are complementary to triangulation. Therefore, methods exploiting such visual and contextual cues for depth can be used to provide an additional source of depth information to the structure from motion or multi-view stereo based depth estimation systems.

In this paper, we focus on texture features, geometric context, motion boundary based monocular cues along with co-planarity, connectivity and spatio-temporal consistency constraints to predict depth in videos. We assume that a scene can be decomposed into planes, each with its own planar parameters. We over-segment a video into spatio-temporal regions and compute depth cues from each region along with scene structure from geometric contexts. These depth cues are used to train and predict depth from features. However, such appearance to depth mappings are typically noisy and ambiguous. We incorporate the independent features to depth mapping of each spatio-temporal region within in a MRF framework that encodes constraints from scene layout properties of co-planarity, connectivity and occlusions. To model the connectivity and co-planarity in a scene, we explicitly learn occlusion boundaries in videos. To further remove the inconsistencies from temporal depth prediction, we apply a sliding window to smooth the depth prediction. Our approach doesn't require camera translation or large rigid scene for depth estimation. Moreover, it provides a source of depth information that is largely complementary to triangulation based depth estimation methods~\cite{saxena2007depth}.
\noindent\textbf{The primary contributions of our method to extract depth from videos are:}
\begin{list}{$\bullet$}
{\setlength{\leftmargin}{2ex}\setlength{\labelsep}{.5ex}\setlength{\parsep}{.4ex plus .2ex minus .1ex}\setlength{\itemsep}{0ex plus .1ex}\setlength{\topsep}{0ex plus .1ex}}
\item Adoption of a learning and inference approach that explicitly models appearance to geometry mappings and piecewise scene smoothness;
\item Learning and estimating occlusion boundaries in videos and utilizing these to constrain smoothness across the scene;
\item There is no requirement of a translating camera or a wide-baseline for depth estimation;
\item An algorithm for video depth estimation that is complementary to traditional structure from motion approaches, and that can incorporate these approaches to compute depth estimates for natural scenes;
\item  We present a novel dataset providing ground truth depth for over 38 outdoor videos ($\sim$ 6400 frames). We also undertake a detailed analysis of contribution of monocular features in depth estimation and thorough evaluation on images and videos;
\end{list}
\section{Related Work}
\label{sec:related}

Structure from motion has been widely applied for 3D scene reconstruction from video. Nister \etal ~\cite{nister2005real} have developed and demonstrated 
real-time reconstruction of point clouds and dense 3D maps from videos. Newcombe \etal \cite{newcombe2010live} proposed a structure from motion (SfM) based method to construct a dense 3D depth map from a single camera video. We cannot do justice here to the vast literature on SfM. However, all SfM approaches need translational motion to 
compute depth maps. 

A key differentiator for the class of approaches represented by this paper is that instead of applying projective geometry techniques to the videos, we reconstruct depth maps based on learning the depth structure of general scenes for reconstruction. As a result, our technique can be applied to videos with no translational motion, substantial motion of foreground objects, and also situations where the camera can be still for short durations.

%

Techniques that are most relevant to our work are the semantic scene structure based scene reconstruction works primarily applied to image analysis. Hoiem \etal \cite{hoiem2007occl} proposed a depth ordering of regions in a single image using geometric context and occlusions depth order. Hane \etal \cite{hane2013joint} proposed joint image segmentation and reconstruction by incorporating class-specific geometry prior from training data. Ladicky \etal \cite{ladicky2012joint} performed a joined semantic segmentation and stereo reconstruction in a CRF framework for street view analysis. Liu \etal \cite{liu2009nonparametric} proposed a non-parametric semantic labeling by pixel label transfer given a databased of known pixel labels. Their method was extended to depth transfer in images and videos by Karsch \etal \cite{karsch2012depth}. Depth transfer algorithm is highly dependent on the global similarity of the target image with candidate images. Moreover, this method does not scale (computationally) for large environments as it requires imagery with associated depth for the environment and a test time global image search over the database for each target image of the video. In contrast, our approach for learning depth from local features, and holistic scene semantics is potentially more likely to scale to larger environments and real-time systems.

Saxena \etal \cite{saxena2009make3d} proposed depth estimation from a single image using local image appearance features. Their depth prediction algorithm was improved by Liu \etal \cite{liu2010single}, incorporating semantic class prior for semantic aware image reconstruction. Our work is inspired by these two approaches but differs in the following key respects, First, we develop a formulation for video reconstruction and second, use occlusion boundaries in the reconstruction. Moreover, we take advantage of temporal context in videos by using video segmentation to get spatio-temporal regions and estimate temporal occlusion boundaries.
\vspace{-0.5cm}
\section{Depth Estimation in Videos}
\label{sec:overview}
\noindent We propose an algorithm to estimate the depth in a dynamic video scene. 
Our conjecture is that appearance and depth correlations learned from training data can be combined with video parsing based on motion coherence of patches and occluding boundaries to infer depth of novel videos. If this approach succeeds, then depth can be computed from videos even when moving cameras do not translate
and there are independently moving foregrounds in the scene. We formulate the depth estimation problem as a problem of inference of locally planar structure within image
patches. The problem is formulated within an MRF framework in which learning provides the data term for planar parameters of patches and inter-patch smoothness is modelled as co-planarity that is allowed to be violated at occlusion boundaries. Motion estimation provides the occlusion boundaries. The learned parameters guide the MAP estimation of the unknown scenes which is solved using simultaneous estimation of all parameters.

\begin{figure*}
  \centering
 \includegraphics[width=\textwidth]{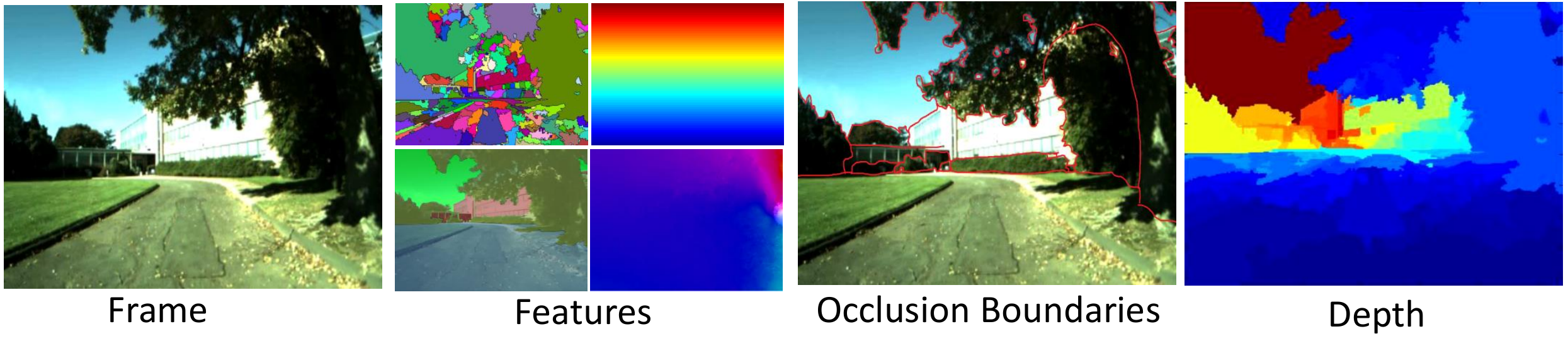}
  \caption{Flow of depth estimation in videos. We first segment the videos into spatio-temporal super-voxels and then extract color, location, texture, motion, and geometric context based features. We predict unary depth prediction with random forest regression, and refine the depth estimate by incorporating 3D scene properties in MRF with occlusion boundaries.}
  \label{fig:overview}
  \vspace{-0.3cm}
\end{figure*}

We process a video in three stages. We first decompose a video into spatio-temporal regions using video segmentation and extract local appearance and motion features for each region (Sec. \ref{sec:vidseg}). In the second stage, we extract the geometric context from the video and apply a random forest based appearance to geometry mapping to predict depth for each spatio-temporal region (Sec. \ref{sec:feats}). Subsequently, we estimate the occlusion boundaries in the video scene (Sec. \ref{sec:occl}) and infer the plane parameters for each spatio-temporal region by enforcing connectivity and co-planarity over non-occluding regions (Sec. \ref{sec:depthMRFfull}). Figure ~\ref{fig:overview} shows the steps of our method. Now we explain each of these steps in detail.

\vspace{-0.3cm}
\subsection{Video Segmentation}
\label{sec:vidseg}
\noindent We assume that a video scene is made up of a small number of planes having uniform depth and orientation \cite{saxena2009make3d,liu2010single}. Therefore, instead of estimating the 3D position of each pixel, we can work on spatio-temporal regions in the video. To decompose a video into spatio-temporal region, we use the graph based video segmentation proposed by Grundmann \etal \cite{MatthiasSegmentation,Corso2012Evaluation}. Their video segmentation algorithm constructs a 3D graph over the video volume and groups together the regions which are coherent in appearance and motion. Their graph based video segmentation processes long sequences of videos by dividing the video into smaller clips and overlapping frames between successive clips to automatically achieve long term coherence. Figure \ref{fig:vidSeg} shows examples of spatio-temporal regions produced by video segmentation \cite{MatthiasSegmentation} \footnote{using www.videosegmentation.com}.

\vspace{-0.4cm}
\subsection{Features:}
\label{sec:feats}
\vspace{-0.2cm}
\noindent Based on the observation that appearance of surface patches in the natural world varies with depth in an imaged scene, we compute a number of appearance and motion based features for each spatio-temporal region obtained by the video segmentation. These features are used to learn a mapping from features to depth as described below. 
\vspace{-0.3cm}
\paragraph{Appearance:}
To capture the appearance of a segment, we compute color, 2D location, and texture features for the segment. Color of a segment is represented by the mean RGB and HSV values within the segment. We apply the texture filters proposed by Leung \etal \cite{LMTexture} and compute filter responses binned into $15$ bins. We assume that camera is approximately parallel (horizontal) to the ground, therefore, locations of vertical regions can provide useful information about the depth. We compute the mean location of normalized y-axis and the vertical distance from the horizon as location features for a segment.
\vspace{-0.3cm}
\paragraph{Motion:}
Motion provides useful information for depth in the scene, \eg, segments closer to the camera exhibit more motion than the segments far from the camera. To include the motion features in our framework, we compute an $8$-bin histogram of dense optical flow \cite{wedel2009improved}, and mean flow within each segment. We also compute the histogram of flow along $x$ and $y$ derivatives. To account for different spatial scales, we compute derivatives for different kernel sizes of Sobel filters ($3$, $5$ and $7$). In addition, we compute the optical flow histograms from reference frame $I_j$ to $I_{j-1}$, $I_{j-3}$, and $I_{j-5}$ to account for temporal smoothing.
\vspace{-0.3cm}
\paragraph{Geometric Context:}
\label{sec:geometric}
Understanding the geometric context of a scene can provide useful information for depth estimation, \eg sky is farthest and ground is horizontal. An object at a distance will have less motion than the object closer to the camera. Moreover, local appearance and motion features are dependent on the geometric class, \eg, appearance of a tree and object may vary differently with distance than appearance of a building, therefore, we include geometric layout of each region in our algorithm. Specifically, we estimate the geometric context in a video scene by the method proposed by Raza \etal \cite{Raza2013GCFV}, which extends \cite{HoiemGeometric2005} from images to videos. They decompose a video into geometric classes given as sky, ground, solid, porous, and moveable objects. Their geometric context from video algorithm predicts the confidence for each pixel belonging to a geometric class. We compute the mean confidence of all the pixels in a spatio-temporal region produced by the video segmentation and include it as a feature vector to the appearance and motion features. 
\subsection{Features to Depth Mapping}
\label{sec:unary_depth}
We use random forest regression to estimate the depth for each 2D region in a frame. Random forests provide feature selection and estimate the out-of-the-bag feature importance \cite{breiman2001random}. Moreover, random forests have been shown to perform excellent over large datasets with huge number of variables. We train random forests with $105$ trees, $11$ random features per node, and a maximum depth of $35$ nodes for a tree. In training random forests, we include all the temporal variations of a 2D segment within its spatio-temporal region provided by the video segmentation. We train and test the random forests over $log$ depth rather than depth to model the depth of nearby regions more accurately.
\vspace{-0.3cm}
\subsection{MRF Framework}
\label{sec:depthMRFfull}
The region-based depth predictions obtained from the random forest learning are typically noisy and do not exploit the piecewise regularities and smoothness structure inherent in most natural scenes. In particular, depths across segments change smoothly except at occluding boundaries. We co-exploit segment-based depth estimation with cross-segment smoothing within an MRF formulation.
A scene in the video is composed of spatio-temporal regions. We model depths within a segment using planar patches and model cross-segment smoothness using 3D connectivity and co-planarity constraint. The cross-segment constraints are allowed to be violated based on evidence from an occluding boundary.
We assume a pinhole camera with no lens distortion. A 3D point in world coordinate is projected onto an image plane pixel $p$. The ray $r_p$ in the world passing through pixel $p$ is given as $r_p=R^{-1}K^{-1}\begin{bmatrix}u_p & v_p &  1\end{bmatrix}^T$, where $(u_p,v_p)$ are image coordinates of pixel $p$. For simplicity, we assume that the camera is parallel to the ground with $\theta_{yz}=0$, and rotation matrix is identity. Note that any of the well-known methods for camera motion estimation from videos can be applied in this context to derive an instantaneous rotation and translation with respect to a reference frame. This can be easily incorporated within our framework. However, since our focus in this work is to explore dynamic depth estimation, we simplify the camera pose problem for expediency. $K$ is the camera matrix, given as
$$
K=\begin{bmatrix}
f_u & 0 & u_o\\
0 & f_v & v_o\\
0 & 0 & 1\\
\end{bmatrix}
$$
where $f_u$ and $f_v$ are the camera focal lengths and $u_o$ and $v_o$ is the image center.

We make the commonly used assumption that each spatio-temporal segment within a video can be described as a planar patch. So all the pixels $p$ in a segment $S$ can be parametrized by plane parameters $\alpha \in \mathbf{R}^3$. The orientation of the plane is given as $\hat{\alpha}=\frac{\alpha}{|\alpha|}$ and the distance of a pixel $p$ on the plane with parameter $\alpha$ is given as $d_p=1/{r_p^T\alpha}$.

The goal of inference in our approach is to estimate $\alpha$'s for each segment in the video within an MRF framework that combines unary terms computed from the appearance to depth mapping, and pairwise terms that enforce connectivity, and co-planarity between planes conditioned on the occlusion boundary between them. The MRF formulation is given as:
\begin{equation}
\label{eq:depth}
P(\alpha|X,S,y)=\frac{1}{Z}\prod\limits_{i=1}^Ng_i(\alpha_i|X_i,S_i)\\
\prod\limits_{ij}f_{ij}(\alpha_i,\alpha_j,y_{ij})
\end{equation}
Where for segment $i$, $X_i$ is the vector of appearance and motion features, $S_i$ is the segment ID, and $y_{ij}$ is the probability of a boundary to be a non-occluding boundary between segments $i$ and $j$. The term $f_{ij}$ is the potential function that measures violations of smoothness between connected segments. Therefore, if the boundary between two spatio-temporal regions is an occlusion boundary then $y_{ij}=0$ and the second term $f_{ij}$ in the MRF becomes zero, \textit{i.e.,} both regions are not connected or co-planar. But if the boundary is a  non-occluding boundary between two spatio-temporal regions then $y_{ij}=1$ and connectivity and co-planarity will be enforced. To handle the errors in the occlusion boundary prediction, we include the confidence for a boundary to be occlusion boundary in our MRF framework (Sec. \ref{sec:occl}). 

The term $g_i(.)$ in Equation 1 minimizes the total fractional depth error over all the pixels $k$ in a region $i$ in a frame. The fractional error is given as
\begin{equation}
\frac{\hat{d}_{k,i}-d_{k,i}}{d_{k,i}}=\hat{d}_{k,i} \cdot r_{k,i}^T\alpha_i-1
\end{equation}
Where $d$ is the ground truth depth and $\hat{d}$ is the predicted depth by the random forest mapping (Sec. \ref{sec:unary_depth}). The data term to minimize the total fractional error in a region $i$ with plane parameter $\alpha_i$ is given by:
\begin{equation}
g_i(\alpha_i|X_i,S_i)=\exp(\sum\limits_{k=1}^K||(r_{i,k}^T\alpha_i)\hat{d}_{i,k}-1||^2)
\end{equation}
Connectivity constrains the pixels on the boundary to be connected by minimizing the fractional distance between the pixels on the boundary $B_{ij}$ of regions $i$ and $j$ in absence of the occlusion boundary. The connectivity term is given as:
\begin{equation}
f_{\mbox{conn}}(\alpha_i,\alpha_j,y_{ij})=\exp(\frac{1}{|B_{ij}|}\sum\limits_{p\in B_{ij}}y_{ij}\\||(r_p^T\alpha_i-r_p^T\alpha_j)\sqrt{\hat{d_i}\hat{d_j}}||^2)
\end{equation}
where $B_{ij}$ are the pixels on the boundary of regions $i$ and $j$ in a frame. The co-planarity term minimizes the fractional distance between the centers of the region and is given by:
\begin{equation}
f_{\mbox{cop}}(\alpha_i,\alpha_j,y_{ij})=\exp(y_{ij}||(r_q^T\alpha_i-r_q^T\alpha_j)\hat{d}_j||^2)
\end{equation}
where $r_q$ is the center pixel of region $j$.

\subsubsection{Occlusion Boundaries}
\label{sec:occl}
Occlusion boundaries arise in a scene due to depth ordering of the objects and geometric classes. Within and across regions in the scene that are connected by a non-occluding boundary such as an orientation discontinuity or surface markings, depth varies smoothly, while occlusion boundaries also correspond to depth discontinuities. Therefore, we enforce connectivity and co-planarity in the scene conditioned on the probability of occlusion boundaries $y_{ij}$ between the regions $i$ and $j$. To estimate occlusion boundaries in a scene, we compute features for each edgelet in the video. Specifically, we compute the difference in color, geometric context, and motion features for the regions on both sides of an edgelet. In addition, we compute the flow-consistency features along each edgelet boundary \cite{humayun2011learning}.  We use standard pairwise potential term over edgelet occlusion prediction. Our occlusion boundary MRF is given as
\begin{equation} \label{eq:occl}
P(e=\mbox{n-occl}|X)=\frac{1}{Z}\prod\limits_{n=1}^Ng_n(e_n|X_n)\\
\prod\limits_{m\in \mbox{conn.(n)}}f_{mn}(e_n,e_m)
\end{equation} 
The term $g_n(.)$ captures the probability for an edge being a non-occlusion boundary given the edgelet feature $X_n$. We compute the unary term by training a random forest with the ground truth from the video geometric context dataset \cite{Raza2013GCFV}. The boundaries between different objects and geometric classes, \ie, sky, ground, trees, building, and objects provide occlusion boundaries and other boundaries correspond to non-occlusion boundaries. The pairwise term $f_{mn}$ is $\sqrt{P(e_n)P(e_m)}$, where edgelet $m$ is connected to edgelet $n$. Next, to make our occlusion boundary prediction temporally consistent, we smooth the occlusion boundary probabilities $P(e=n-occl|data)$ over a temporal window of $30$ frames. Note that the edgelet $e_n$ in Equation 2 is the boundary between regions $i$ and $j$ in Equation 1, and occlusion probability of an edge $P(e_n=n-occl|data)$ is the same as $y_{ij}$. 

Now, we solve the \Cref{eq:depth} using the L-BFGS algorithm \cite{liu1989limited} and compute the depth of each pixel $p$ in region $i$ using $d_{i,k}=1/r_{i,k}^T\alpha_i$. We compute depth for each 2D region in a frame using the MRF and then use a sliding window across frames to smooth out the inconsistencies in predicted depth.
\section{Data Collection}
\label{sec:data}
\paragraph{Existing Datasets:}
We propose an algorithm to extract temporal depth information in outdoor videos using scene structure (geometric context and occlusion boundaries). For training and evaluating our algorithm, we require a video dataset with depth ground truth. Unfortunately, such a dataset is not available. NYU RGB-D dataset provides depth ground truth for indoor scenes collected with Kinect sensor \cite{Silberman:ECCV12}. Karsch \etal \cite{karsch2012depth} developed a video ground truth dataset for indoor videos using Kinect but Kinect cannot produce ground truth for outdoor videos due to IR interference. KITTI dataset provides outdoor videos with depth ground truth from velodyne \cite{Geiger2013IJRR}. However, the velodyne is mounted on a car to maximize the depth accuracy along the road, so most of the trees, buildings and other scene structures don't have any depth information. Since our study is focused on extracting the depth in videos combined with scene description (geometric context, and occlusion boundaries), missing ground truth depth of the whole scene makes KITTI dataset unsuitable for our study. The only existing dataset with full scene ground truth depth is Make3d dataset\cite{saxena2009make3d}, which is only limited to images. Therefore, for depth estimation in videos, we have developed a video dataset for ground truth depth.
\paragraph{Video Depth Dataset for Scene Understanding:}
We collected a video depth data set from a robot mounted by an RGB camera and a velodyne HDL-32E rotating 3D laser scanner. The HDL-32E velodyne has 32 laser/detector pairs, $+10.67$ to $-30.67$ degrees vertical field of view at $10$Hz frame-rate. It results in a total of $700K$ laser points per second over a range of $80$m. We estimate the velodyne to camera transformation by the method proposed by Unnikrishnan \etal \cite{unnikrishnan2005fast}. Now we can project a 3D points $\mathbf{x}$ from velodyne coordinates to image coordinates $\mathbf{y}$ using
\begin{equation} \label{eq:projection}
\mathbf{y}= \mathbf{K} \mathbf{T_{velo}^{cam}} \mathbf{x}
\end{equation}
where $K$ is intrinsic matrix. $\mathbf{T_{velo}^{cam}}$ is the transformation matrix from velodyne coordinates to camera coordinates given as
$$
\mathbf{T_{velo}^{cam}}=\begin{bmatrix}
\mathbf{R_{velo}^{cam}} & \mathbf{t_{velo}^{cam}} \\
0 & 1\\
\end{bmatrix}
$$
where $\mathbf{R_{velo}^{cam}}$ and $\mathbf{t_{velo}^{cam}}$ are rotation and translation matrices from velodyne to camera.

Using the transformation given in Equation 8, we project velodyne laser points on a video. However, since velodyne laser points have lower resolution than video, we leverage from video segmentation to generate cleaned depth maps. We compute the average depth of all the laser points project on a spatio-temporal region, over a temporal window of 5 frames to generate the ground truth depth.
\section{Experiments and Results}
\label{sec:result}
\begin{figure*}
  \centering
      \includegraphics[width=0.75\textwidth]{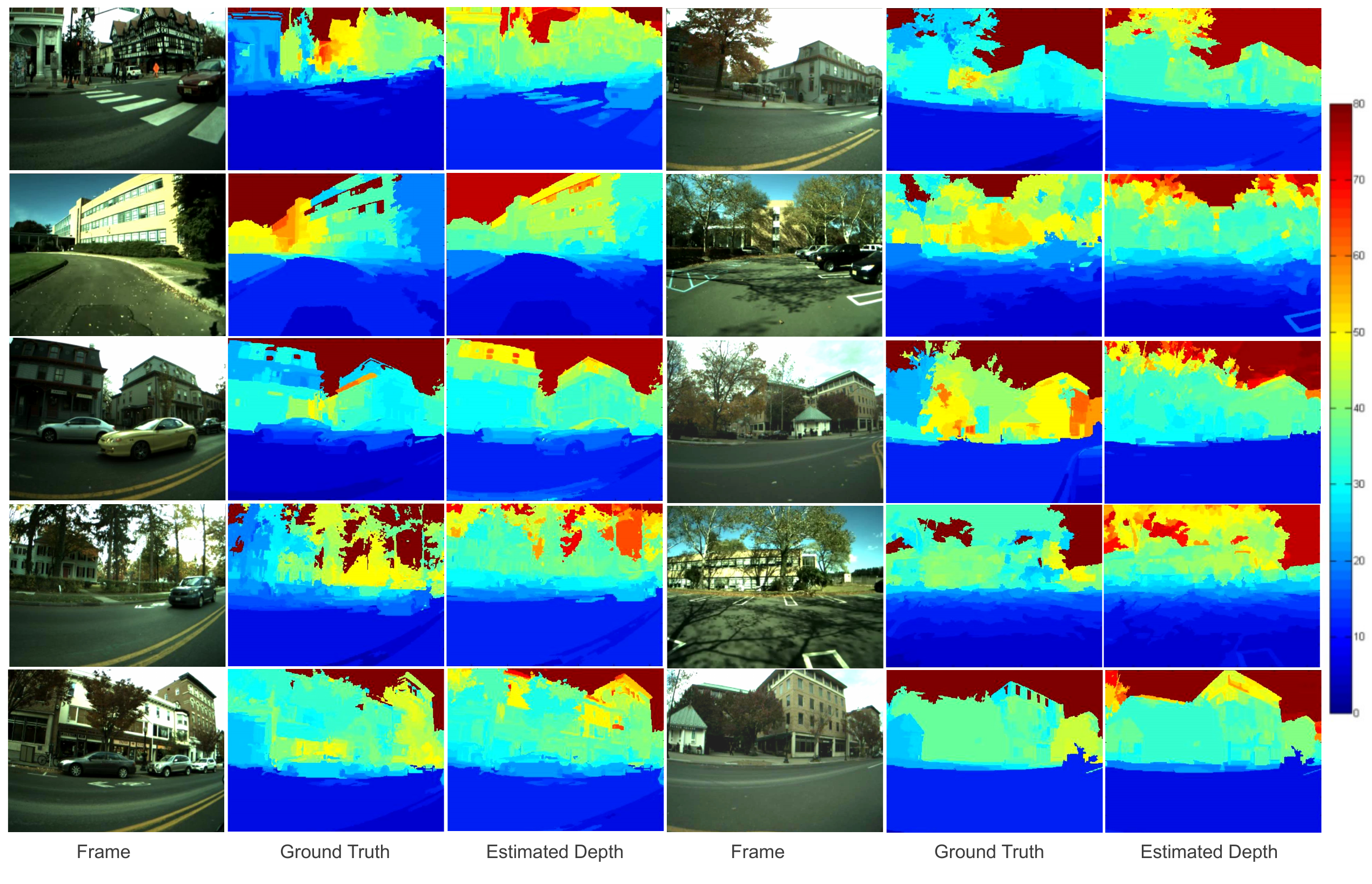}
  \caption{Examples of videos scenes, ground truth, and predicted depth by our method. Legend shows depth range from 0m (blue) to 80m (red). }
  \label{fig:vid_depth_result}
\vspace{-0.5cm}
\end{figure*}

\noindent We perform extensive experiments on video depth data to evaluate our algorithm. We perform 5-fold cross-validation over 36 videos ($\sim$ 6400 frames). We compute average log-error $|\log d - \log {\hat{d}}|$ and average relative error $|\frac{ d - \hat{d}}{d}|$ to report the accuracy of our method. We achieve an accuracy of $0.153$ log-error and $0.44$ on relative error (Table \ref{Table:result}). Figure \ref{fig:vid_depth_result} shows some example scenes from our dataset with ground truth and predicted depth.

\begin{figure}
\begin{minipage}[t]{0.47\linewidth}
  \centering
      \includegraphics[width=0.8\columnwidth]{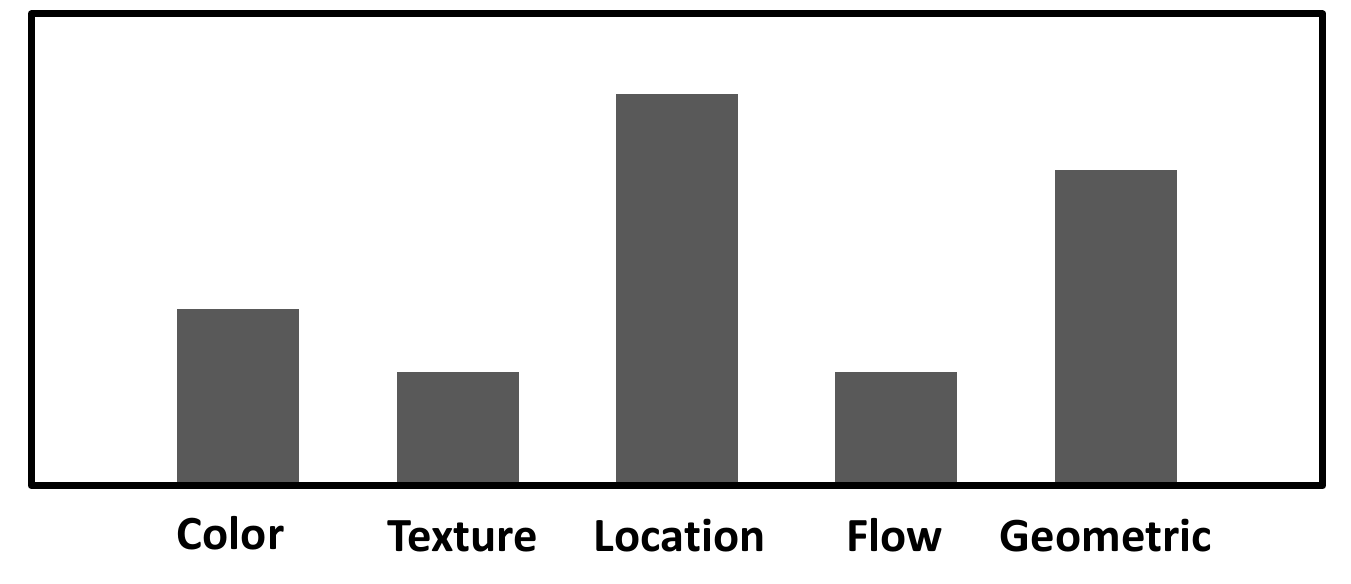}
  \caption{Random forest's out-of-bag features importance for depth regression.} 
  \label{fig:feat_imp}
\end{minipage}
\hspace{0.5cm}
\begin{minipage}[t]{0.47\linewidth}
  \centering
      \includegraphics[width=0.8\textwidth]{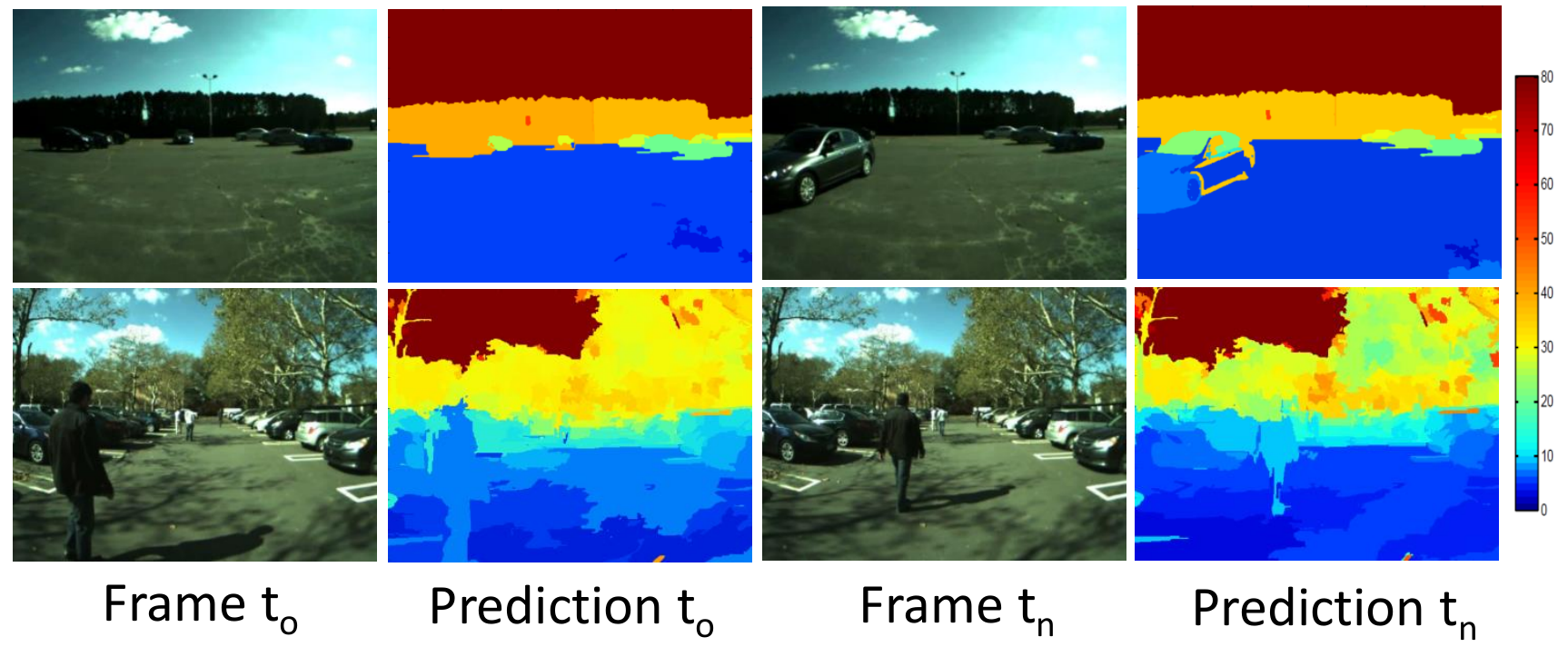}
  \caption{Our method can predict temporal object depths with high accuracy.}
  \label{fig:obj_depth}
\end{minipage}
\end{figure}

We also compute depth accuracy over each predicted geometric class. We achieve good performance for objects, ground, and sky. Geometric context provides high accuracy for ground and sky prediction, contributing to better depth prediction for these classes. We achieve comparatively lower accuracy for trees and buildings. Highly porous tree branches and glass buildings can have error in ground truth depth due to velodyne limitation. Moreover, occlusion boundary prediction within porous regions can have lower accuracy. However, using geometric and motion features enable our algorithm to accurately predict temporal depth for moving objects, \eg, Figure \ref{fig:obj_depth} shows examples of temporal update in object depth prediction.


\Cref{Table:result} shows the importance of features for depth estimation on video depth dataset. It shows that geometric features provide useful information for depth estimation. Figure \ref{fig:feat_imp} shows average out-of-bag feature importance by random forest regression over 5-fold cross-validation normalized by sub-feature dimension. Figure \ref{fig:feat_imp} shows that location and geometric features provide most useful information for the random forest. Motion features also provide useful features but have lower average feature importance because of high feature dimension. Motion features are also useful for extracting geometric context in videos and occlusion boundaries detection.

Our approach for depth estimation can also be applied to images. We applied our algorithm over a publicly available Make3d depth image dataset \cite{saxena2009make3d}. The Make3d dataset contains $400$ training images and $134$ test images. To apply our video-based algorithm to images, we first apply image segmentation by Felzenshawb \etal \cite{superpixel}. Next, to estimate geometric context, we apply the publicly available code from Hoiem \etal \cite{HoiemIJCV2007}. For occlusion boundary prediction, we train our random forest classifier with only the appearance based features from video geometric context dataset \cite{Raza2013GCFV}. Table \ref{table:make3d} gives the comparison of the single image variant of our approach with the state of the art and we achieve competitive results. It should be noted that our algorithm depends on occlusion boundary detection and geometric context (for which motion based features are important \cite{stein2009occlusion,Raza2013GCFV}) and is not optimized to extract depth from single images.

\begin{table}

\begin{minipage}[t]{0.47\linewidth}
\centering
\small
\begin{tabular}{|c|c|c|c|}\hline
Features & $log10$  & rel-depth \\\hline
ALL & 0.153  & 0.44\\\hline
Appearance+Flow & 0.176  & 0.533\\\hline
Appearance & 0.175  & 0.512\\\hline
\end{tabular}
\vspace{0.1cm}
\caption{Performance of our algorithm on video dataset, combining appearance, flow, and surface layout features give best accuracy.}
\label{Table:result}
\end{minipage}
\hspace{0.5cm}
\begin{minipage}[t]{0.47\linewidth}
\centering
\small
\begin{tabular}{|c|c|c|c|}\hline
Algorithm & $log10$  & rel-$log$ \\\hline
SCN \cite{saxena2005learning} & 0.198  & 0.530\\\hline
HEH \cite{HoiemIJCV2007} & 0.320  & 1.423\\\hline
Baseline \cite{saxena2009make3d} & 0.334  & 0.516\\
PP-MRF \cite{saxena2009make3d} & 0.187  & 0.370\\
Depth Transfer \cite{karsch2012depth} & 0.148 & 0.362  \\\hline
Sematic Labels \cite{liu2010single}  & 0.148 & 0.379 \\\hline
\textbf{*Geom. Context}  & & \\
\textbf{Occl. Bound.} & \textbf{0.159} & \textbf{0.386} \\\hline
\end{tabular}
\vspace{0.1cm}
\caption{Our approach can also be applied to images. We apply it to Make3d depth image dataset \cite{saxena2009make3d}.}
\label{table:make3d}
\end{minipage}
\end{table}

\section{Conclusion}
\label{sec:conclusion}
\noindent Driven by the conjecture that appearance, motion and occluding boundaries in monocular videos afford a depth of the scene, we have proposed a novel algorithm for depth estimation in videos. Our approach does not rely on using the structure-from-motion approaches that require cameras to translate and foreground movements to be minimal. Of course, a promising future work direction is to incorporate robust methods for structure-from-motion for camera and scene estimation within our framework. We have employed learned mappings from appearance and motion of spatio-temporal patches in videos to their corresponding depth estimates using ground truth data collected as co-registered videos and 3D point clouds. These learned mappings are incorporated into an MRF framework in which data terms are derived from these mappings while smoothness across spatio-temporal segments is formulated as the inter-segment constraint. Smoothness is allowed to be violated across occluding boundaries. Evidence for occluding boundaries is computed through edgelets and motion signatures within and across neigboring segments. Quantitative results of our algorithm on static image datasets such as Make3D demonstrate that on static images we are at least as competitive as state-of-the-art approaches. Since there are no standard 3D video datasets, we have shown results on a newly captured datset with co-registered videos and 3D point clouds.
\vspace{-0.5cm}

\section{Acknowledgment}
This work has been supported by the Intelligence Advanced Research Projects Activity (IARPA) via Department of Interior National Business Center contract number D11-PC20066. The U.S. Government is authorized to reproduce and distribute reprints for Governmental purposes not with-standing any copyright annotation thereon. The views and conclusions contained herein are those of the authors and should not be interpreted as necessarily representing the ofﬁcial policies or endorsements, either expressed or implied, of IARPA, DOI/NBC, or the U.S. Government.

\bibliography{egbib}
\end{document}